# Hand-Drawn Electrical Circuit Recognition using Object Detection and Node Recognition

**Rachala Rohith Reddy [1] and Mahesh Raveendranatha Panicker \***

## Abstract

With the recent developments in neural networks, there has been a resurgence in algorithms for the automatic generation of simulation ready electronic circuits from hand-drawn circuits. However, most of the approaches in literature were confined to classify different types of electrical components and only a few of those methods have shown a way to rebuild the circuit schematic from the scanned image, which is extremely important for further automation of netlist generation. This paper proposes a real-time algorithm for the automatic recognition of hand-drawn electrical circuits based on object detection and circuit node recognition. The proposed approach employs You Only Look Once version 5 (YOLOv5) for detection of circuit components and a novel Hough transform based approach for node recognition. Using YOLOv5 object detection algorithm, a mean average precision (mAP$_{0.5}$) of 98.2% is achieved in detecting the components. The proposed method is also able to rebuild the circuit schematic with 80% accuracy with a near-real time performance of 0.33s per schematic generation.

---

✉ [1] R Rohith Reddy

rohithreddy0087@gmail.com

Department of Electrical Engineering,

Institute of Technology Palakkad, Kerala 678623,

India.

\* Mahesh Raveendranatha Panicker

Corresponding Author

mahesh@iitpkd.ac.in

Department of Electrical Engineering,

Institute of Technology Palakkad, Kerala 678623,

India.

## 1. Introduction

While designing electronic circuits, the initial circuit ideas will be typically drawn on a paper and later for simulation purposes the circuit will be redrawn using a simulation software. Even though, there are many research works which are proposed to efficiently digitalize texts and scripts, very few studies have been proposed to digitize electronic circuits. Hence, an automatic conversion of hand drawn electronic circuit to a schematic diagram readily usable with simulation programs such as LT Spice or PSpice will be extremely beneficial.

Digitizing an electrical circuit involves efficient detection of components and accurate tracing of connections between them. Since the target is handwritten, the different variations in drawing style make the process of recognition much harder. Different variations in the quality of paper and ink, the noise acquired while capturing the image are some major challenges in the field of circuit recognition. A model which is robust to these different variations is needed to make the process of circuit recognition more effective.

Conventional image processing techniques have been applied but they weren't effective enough to accurately rebuild the image, hence in our work we employ deep learning techniques that have been proved to be robust and efficient to overcome the challenges. In this work, the popular and light weight YOLOv5 neural network architecture is employed to detect the components and a Hough transform based approach is employed for node recognition. The rest of the paper is organized as follows. A comprehensive literature review is presented in section 2. The proposed framework is explained in section 3. The results along with the dataset are presented in section 4 followed by conclusions.

## 2. Literature Review

Although hand-drawn circuit recognition is not a well-researched area, there are some studies that investigated this area. Most of these research works focus only on the classification of different types of circuit components like resistors, capacitors etc. and only a few describe a method to trace the connections/wires between these components. In [1], a two-stage convolutional neural network (CNN) network is proposed to classify 20 different components, which has been trained on 1050 images (original + augmented) per class. Circuit components with similar shape and structure are grouped together in the first stage and are then classified into actual classes using CNN models, which are specific to each group in the later stage. Using this model, a recognition accuracy of 97.33% is achieved. Roy et al. [2] have proposed a recognition model which is based on a feature set containing of histogram of oriented gradient (HOG) descriptor (based on texture features), and features derived from shape such as chain code histogram, tangent angle and centroid distance. Using a feature selection algorithm called ReliefF, texture-based features are optimized and then are classified using a sequential minimal optimization (SMO) classifier. This classifier achieves an accuracy of 93.63%. Using features such as area, centroid and eccentricity, a k-nearest neighborhood (KNN) based approach [3] is used to classify the electrical components from the image using the image's moments. This approach achieves an accuracy of around 90% in classification. In [4], the HOG features are employed along with support vector machine (SVM) classifier to classify the components. However, the accuracy of classification is not mentioned. The approach in [5] considers some geometric properties like a rectangular box, curvature, slope, width, and number of intersects to extract the features, these features are then applied to the Hidden Markov Model (HMM) classifier. This approach classifies the connectors or components with a recognition accuracy of 83%. An Artificial Neural Network (ANN) based approach is employed in [6] to classify the different electrical components using shape-based features extracted from electrical circuit image. Using these features a ANN is trained, although the method is viable, accuracy obtained is much lower. Another HOG + SVM based approach is presented in [7], where before the recognition of components, the circuit image is segmented using morphological operations, these segmented components are then used to classify. A segmentation accuracy of 87.70% and a classification rate of 92.00% is achieved using this method. A two-dimensional dynamic programming-based technique is presented in [8], that employs symbol hypothesis generation, which can efficiently segment and recognize interspersed symbols. Using this method an accuracy of more than 90% is achieved in recognizing electrical components. All the above-related works only propose a method to classify the components in the circuit. But to completely recognize a circuit, it is important to detect and trace both components and connections in the circuit. Some of the works which described a method for circuit schematic generation from hand drawn circuits are briefly explained below.

In [9], an approach is presented, in which the scanned circuit image is first converted into a gray scale image and then to a binary image by thresholding the image. Later, any small spots or noise are removed from image by using suitable image processing techniques and then converted to a single-pixel line image using morphological operations. Using thresholding techniques on spatially varying pixel density, nodes and components are segmented, then using pixel stack connection paths are extracted. Syntactic analysis is used to classify nodes. Classification of the components is done using scalar pixel distribution features, a combination of invariant moments and vector relationships between straight lines and polygons. This method achieves an accuracy of 92% in node recognition and accuracy of 86% in component recognition, on a database comprising of 107 nodes and 449 components. In [10], an entirely different perspective is presented, where instead of detecting the components, the unit of the components such as Ω (ohms), F (Farad), H (Henry) is identified using well established optical character recognition techniques. To extract the wires, all the jagged lines (e. g. resistor) in the circuit are removed and the Hough transform is applied to get both horizontal and vertical lines. Then the nodes in the circuit are identified as the terminals of the components, and these are mapped to the centroids of the characters recognized. This method doesn't work if there are no units written on the circuit, but the method described to extract the connections is noteworthy.

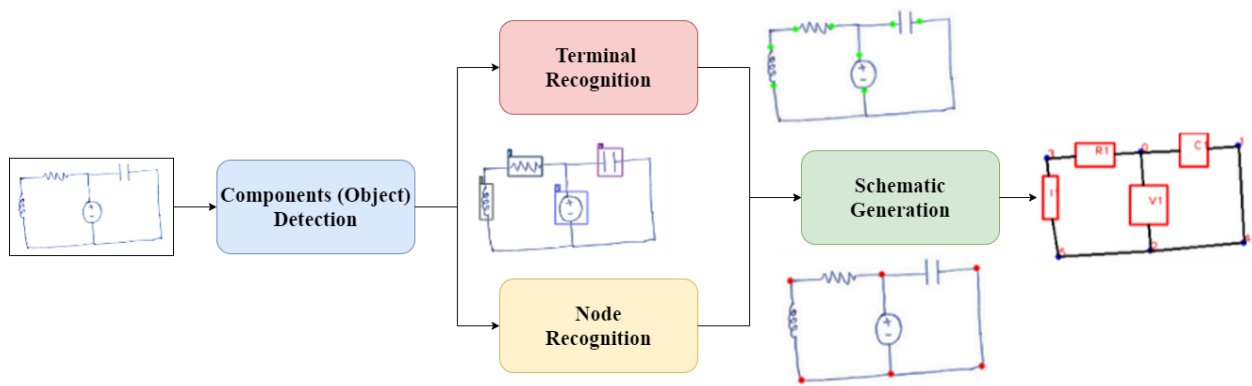

**Fig. 1** Proposed framework for circuit component detection, node and terminal recognition and schematic generation.

The methods [1-8] are only able to classify the components in the circuit whereas the method in [9] can classify the components and is also able to trace the connections in the circuit. The classifier used in [9] does not give good accuracy and the whole process is dependent on thinning the circuit which is performed before component classification and node recognition. Most of the scanned images, when passed through a thinning morphological operation, lose the data i.e., the image gets eroded, and gaps are created between the lines. Hence, the thinned image cannot be used to trace the connections. Although the method [10] has a good approach to extract the wires, they lack an efficient method to detect the components.

To summarize, there were approaches that were proposed in the literature towards circuit component classification, there have been hardly any work towards an end-to-end circuit schematic generation. To the best of our knowledge, the work presented in this approach is the first end-to-end near real-time approach to generate a circuit schematic from hand drawn circuit images with an object detection neural network front end.

## 3. Proposed Method

The proposed approach, as shown in Fig. 1 is an attempt towards near real-time automated circuit schematic generation from the hand drawn circuit images. The approach consists of circuit component detection, node and terminal recognition and schematic generation. In the proposed approach, a lightweight neural network (NN) is employed to detect the components in the circuit. The NN models can be trained to be robust to any changes in the input images, hence the scanned circuit image can be given directly to the model without applying any morphological operation. Also, modern object detection algorithms such as YOLOv5 [11] are very accurate even with training on fewer images and extremely fast, and hence the proposed approach can be used for near real-time applications. The object detection model outputs the bounding boxes and the classes of the components. Using these bounding boxes, the terminals and the nodes of the components can be traced. Using the detected classes, bounding boxes, terminals and nodes, the entire circuit schematic can be generated which can later be mapped to a suitable netlist.

Unlike the approach in [10], the proposed approach assumes that the circuit is present in the scanned image without any component symbols or units. Also as discussed, without any preprocessing of the image, components in the circuit can be detected. A detailed explanation of the building blocks of the proposed approach is given below.

### 3.1 Detection of Circuit Components

One of the key components of the proposed approach is a real-time object detection algorithm. Object detection is a combination of object classification and object localization, using these algorithms any specific object with its bounding boxes can be determined in an image. A convolutional neural network (CNN) based object detection model can be trained to identify and detect more than one specific object, it is versatile. These CNN algorithms use special and unique properties of each class to classify an object. Hence there is no need of using any shape or geometry-based features to identify the circuit components. An ablation study has been done between the three popular object detection NNs YOLOv3, YOLOv5 and single shot detection (SSD) considering the constraints of desirable light weight implementation.

### 1) YOLOv3

You only look once (YOLO) [12] is a state-of-the-art, real-time object detection system. YOLO applies a single neural network to the full image. The neural network divides the image into regions and bounding boxes along with the probabilities for each region are predicted. Using non-maximal suppression (NMS), the boxes that share large areas with other boxes and ones with a low probability of containing an object are removed. YOLOv3 [13] is based on the idea of a residual network (ResNet [14]), it uses a feature extraction network called Darknet-53 [15] and has a total of nine anchor boxes. It has a total of 106 convolutional layers which are activated by leaky rectified linear unit (ReLU) function and followed by batch normalization. While the accuracy of YOLOv3 is high compared to its previous versions, it traded off against the speed making it debatable for real-time applications.

### 2) YOLOv5

YOLOv5 [11] is the latest version in the YOLO series. In YOLOv5, the cross stage partial networks (CSP [16]) are used to extract rich informative features from an input image. YOLOv5 uses PANet [17] to get feature pyramids. In YOLOv5, the leaky ReLU activation function is used in middle/hidden layers and the sigmoid activation function is used in the final detection layer. YoloV5 is by far the fastest and accurate object detection algorithm for natural images.

### 3) SSD300

Single Shot Multibox Detector (SSD) [18] is an object detection algorithm designed for real-time object detection. SSD's architecture has three parts: 1) base convolutions are derived from VGG-16 [19] network which provides low level features, 2) auxiliary convolutions are added on top of the base convolutions to get high level features and 3) prediction convolutions help in locating and classifying objects in these features. There are two models for SSD: 1) SSD300: 300×300 input image, lower resolution, faster and 2) SSD512: 512×512 input image, higher resolution, more accurate. In this study, SSD300 has been employed for circuit component detection.

## 3.2 Terminal Recognition

Once the detection of key circuit components is done, the next step is to trace the connections for generating the schematic. The first step towards tracing of the circuit connections is to identify the terminal points, which are points on either side of the detected components. To extract these terminals, two binary images are generated 1) from the output of the bounding boxes generated by objection detection and 2) adaptive thresholding of the scanned image as shown in Fig. 2. Generally, thresholding is done by choosing a threshold value such that all the pixels with intensities above this threshold are quantized to 1 and below the threshold to be 0 as in (1). The threshold value is taken to be half of the highest intensity as 127 (255/2) in our case assuming uint8 images.

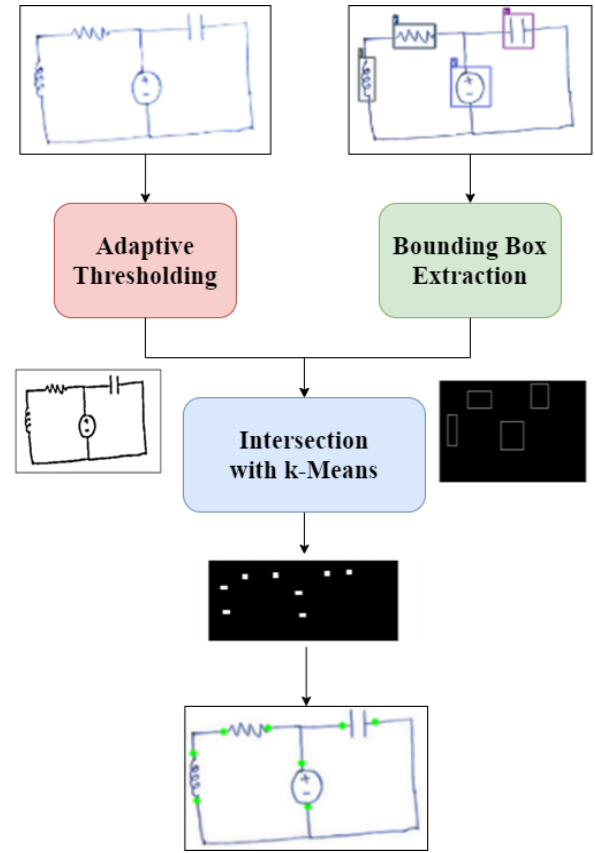

**Fig. 2** Proposed terminal recognition algorithm

$$binary\_image = \begin{cases} 0, & f(x,y) < threshold \\ 1, & f(x,y) > threshold \end{cases} \quad (1)$$

However, the choice of a single threshold might not be good if the image is taken in varying lighting conditions. In such cases, adaptive thresholding works better. The image is divided into different regions and each region has a different threshold value. This threshold value is as given in (2), where the Gaussian-weighted sum of the neighborhood values minus the constant C in that region.

$$binary\_image\,(w,w) = \begin{cases} 0, & f(w,w) < (\sum_x \sum_y G(w,w)f(w,w)) - C \\ 1, & f(w,w) > (\sum_x \sum_y G(w,w)f(w,w)) - C \end{cases}$$
(2)

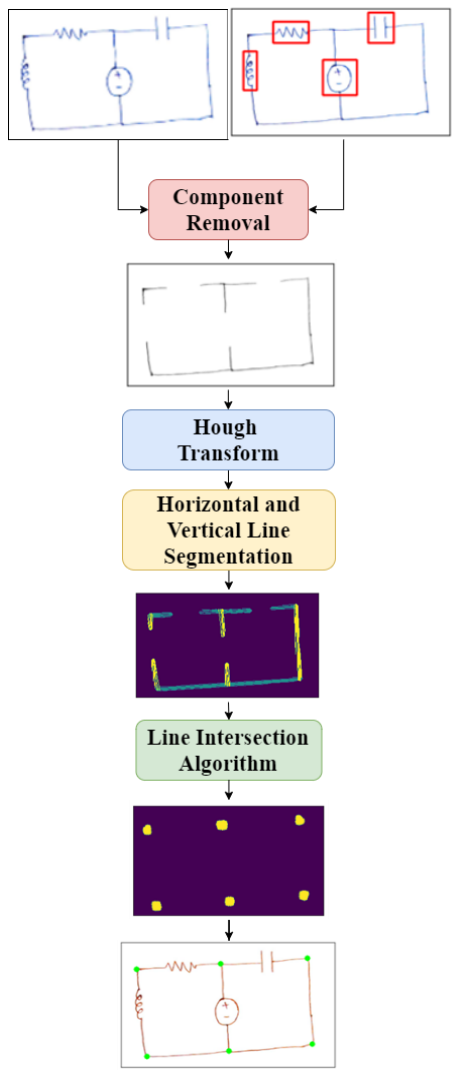

**Fig. 3** Proposed node recognition algorithm

Where $G(w, w)$ is a Gaussian window of size $w \times w$ and $C$ is a constant. Based on the intersection of the adaptive thresholded binarized circuit and the thresholded and binarized bounding boxes, the terminals can be identified. However, to improve the results, the centre of the intersections can be estimated using k-means clustering which results in a robust estimate of the terminal points. To perform k-means clustering one needs to know the number of centroids. The number of centroids, in this case, are twice the number of components obtained from the object detection algorithm. With randomly selected points as centroids, an iterative computation is performed to optimize the position of the centroids.

### 3.3 Node Recognition

Once the terminals are obtained, the next step in the framework is to identify the nodes in the circuit. Towards this, an image with only the connections will suffice. This means all the components in the given input scanned image should be removed. Since bounding boxes for each component in the given circuit are obtained as in section 3.1, the coordinates of these bounding boxes can be employed in such a way that all the area inside these bounding boxes can be made as white pixels as illustrated in Fig. 3 (output of the component removal block).

In the image of connections separated from the circuit diagram as in Fig. 3, nodes can be considered as the intersection points of horizontal and vertical lines. Towards this, the next step is to employ Hough Transform to detect the lines. Then these lines are segmented based on the orientation of the line as mostly horizontal and vertical lines will be of interest to us. Since, all the line segments are obtained from the Hough transform, the starting and ending points of these line segments are also obtained. Hence, from these points, the slope of the line can be determined, if the slope of the line is greater than $45^o$ and less than $135^o$, it is considered as a vertical line and if the slope of the line is less than $45^o$ and greater than $135^o$, it is a horizontal line as in (3-4).

$$slope = \frac{y_2 - y_1}{x_2 - x_1} \tag{3}$$
$$where\ (x_1, y_1)\ and\ (x_2, y_2)\ are\ ends\ of\ the\ line\ segment$$

$$type\ of\ line =$$
$$\begin{cases} horizontal\ line,\ 45^0 < slope < 135^0 \\ vertical\ line,\ slope > 135^0 and\ slope < 45^0 \end{cases} \tag{4}$$

Using the above condition, horizontal and vertical lines can be segmented (shown as yellow and blue coloured lines for vertical and horizontal lines in Fig. 3). Once the horizontal and vertical lines are identified, nodes can be estimated as the intersection points of these lines. The intersection of lines algorithm can be used to find out the intersection point and the location of the intersection point between the start and terminal of these lines can be employed for classifying it as a node or not. Suppose following are the two different line equations that are formed by $\{(x_{11}, y_{11}); (x_{12}, y_{12})\}$ and $\{(x_{21}, y_{21}); (x_{22}, y_{22})\}$ respectively:

$$Line\ 1 => a_1 x + b_1 y = c_1$$
$$where\ a_1 = y_{12} - y_{11}, b_1 = x_{11} - x_{12}, c_1 = a_1 x_{11} + b_1 y_{11}$$

$$Line\ 2 => a_2 x + b_2 y = c_2$$
$$where\ a_2 = y_{22} - y_{21}, b_2 = x_{21} - x_{22}, c_2 = a_2 x_{21} + b_2 y_{21}$$

Then the point of intersection between these two lines is as follows (5):

$$(x, y) = \left( \frac{b_2 c_1 - b_1 c_2}{a_1 b_2 - a_2 b_1}, \frac{a_1 c_2 - a_2 c_1}{a_1 b_2 - a_2 b_1} \right) \tag{5}$$

Since we are finding the point of intersections between two different line segments, these points should always lie inside these line segments i.e. (6-7),

```
# k number of terminals
terminals = [t1,t2,.....,tk]
# m number of nodes
nodes = [n1,n2,......,nm]
# declare a dictionary with terminals as keys and values as nodes
terminal_node_map = dict
for t in terminals:
    # choosing a maximum value
    min_dist = 9999
    arg_min = 0
    for i in range(nodes):
    # measuring euclidean distance
        dist = distance(n[i],t)
            if dist < min_dist:
                min_dist = dist
                arg_min = i
    terminal_node_map[t] = arg_min
```

**Fig. 4** Pseudo code to map the nodes and terminals

$$\min(x_{11}, x_{12}, x_{21}, x_{22}) \leq x \leq \max(x_{11}, x_{12}, x_{21}, x_{22}) \quad (6)$$

$$\min(y_{11}, y_{12}, y_{21}, y_{22}) \leq y \leq \max(y_{11}, y_{12}, y_{21}, y_{22}) \quad (7)$$

If the point of intersection doesn't satisfy the above condition, then it is not considered for node detection. This is because, the point of intersection may sometimes lie outside the lines and instead of getting the required number of intersections, more intersections are detected which leads to the detection of a greater number of nodes than that are originally present. This fails the process of circuit recognition and hence above criteria must be met to consider a point of intersection for node detection. Before finding the number of nodes, the image is dilated to cover all the small gaps present in between the points of intersection otherwise a greater number of nodes may be detected (the dilated points are shown as the output of the line intersection algorithm and not the actual distribution of points). The number of nodes in the circuit is typically equal to the number of contours. Contour is an outline bounding the shape of a node, hence using contour detection on the dilated intersection points will give the number of nodes. In most cases, contour detection gives the exact number of nodes in the circuit. But the above-obtained nodes are strictly not localized points, instead, they are regions, hence, to find the exact coordinates of these nodes, k-means clustering can be used as described in section 3.2. The estimated nodes overlaid on the scanned image is as shown in Fig. 3 (final image).

### 3.4 Circuit Schematic Generation

In most of the electrical circuits, the terminals of a component are connected to the nearest node, but this might not be true in all cases, taking account of the majority cases, a distance-based algorithm is proposed. After obtaining all the terminals, nodes, and components in the circuit, connections can be traced by mapping them to each other. Under the assumption that in a circuit, each terminal is connected to a node, and hence the connections can be made by choosing the minimum distance between a particular terminal and all the nodes in the circuit, a novel approach is presented to map the nodes to the terminals. In general, each terminal is connected to its nearest node. The pseudo code for proposed method is presented in Fig. 4. After mapping all the terminals to their respective nodes, the nodes which are connected to less than two terminals are connected to each other. A similar method as above presented pseudo code is followed to map the nodes to each other. This way all the connections are made and the process of rebuilding the circuit is completed.

## 4. Results and Evaluation

### 4.1 Dataset

Since there is no publicly available dataset, a custom dataset is generated by collecting hand-drawn circuits drawn by five different people. This paper focuses only on certain electrical components such as voltage source, resistor, capacitor, inductor, and diode, however the algorithm could be extended to any component with necessary retraining. Hence the training set consists of electrical circuits only with these components as shown in Fig. 5. A total of 154 circuit images are collected and manually labelled and out of these 154 images, 103 images are used for training the NN model and 51 images are used for validating and testing the model. To keep the categories balanced, almost all classes are equally represented. To increase the data for training, data augmentation like rotating and flipping the image are used to create a dataset with 388 images for training.

### 4.2 Training Methodology

Using PyTorch framework and the YOLOv5, YOLOv3 and SSD300 networks as a model, component detection is trained. The images are resized as 416x416 for YOLO and 3030x300 for SSD models respectively. The other parameters for training the YOLO models are as follows:

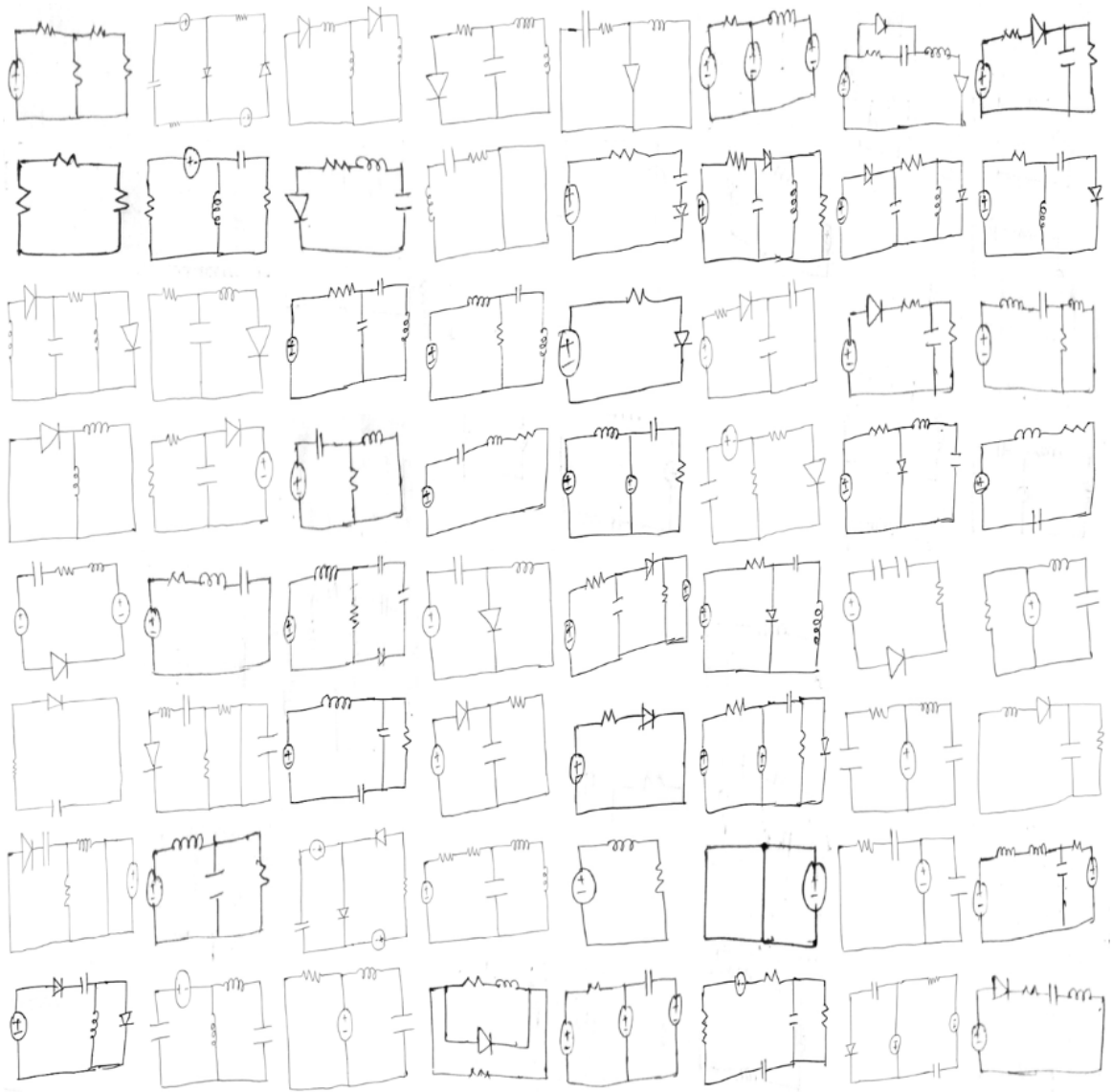

**Fig. 5** A sample dataset containing scanned images of hand drawn electrical circuits used for training of object detection algorithms

learning rate was 0.001, momentum was 0.937, the batch size was set to 16 and the number of epochs for which the model was trained is 500. The parameters for training SSD model are as follows: learning rate was 0.001, momentum was 0.9, batch size was set to 8 and the number epochs were 1500. These parameters are similar to that of training a YOLOv5 model on COCO dataset (Common Objects in Context) as in [12].

### 4.3 Evaluation Method

In object detection, there are two distinct tasks: one is determining whether an object exists in the image, and the other is to find the object's location. In this paper the mAP (mean Average Precision) was used as the evaluation criteria. The mAP score is determined by calculating the AP over all classes and all Intersection over Union (IoU) thresholds. In this experiment, the IoU threshold was considered to be 0.5 and the mAP was averaged over the five object classes.

To evaluate the performance of the classification by the object detection algorithms, metrics like accuracy, precision, recall and F1-score are computed. These are calculated from the confusion matrix. Using the following steps confusion matrix can be computed for an object detection algorithm.

1. For each detection, the algorithm extracts detected boxes, classes, scores and along with these ground-truth boxes, classes are present. Only detections with a score greater than or equal to 0.5 are considered. Anything that's under this value is discarded.

2. For each ground-truth box, a match is made if the IoU between the ground-truth box and detected box is greater or equal than 0.5.

3. The duplicates in the list are removed (ground-truth boxes that match with more than one

| Backbone | mAP$_{0.5}$ (%) | AP (Voltage source) (%) | AP (Capacitor) (%) | AP (diode) (%) | AP (Inductor) (%) | AP (Resistor) (%) | Time (seconds) |
|---|---|---|---|---|---|---|---|
| YOLOv5 | 98.1 | 100.0 | 100.0 | 99.77 | 90.93 | 99.80 | 0.0270 |
| YOLOv3 | 98.1 | 100.0 | 100.0 | 100.0 | 90.91 | 99.60 | 0.0520 |
| SSD300 | 92.5 | 100.0 | 90.41 | 90.90 | 90.01 | 90.90 | 0.0515 |

**Table 1** Performance Comparison of Object Detection Algorithms

| Class | Accuracy | Precision | Recall | F1-score |
|---|---|---|---|---|
| Voltage Source | 100 | 100 | 100 | 100 |
| Capacitor | 98.34 | 100 | 92.59 | 96.15 |
| Inductor | 98.34 | 97.29 | 92.31 | 94.73 |
| Resistor | 98.34 | 98.15 | 94.64 | 96.36 |
| Diode | 99.17 | 95.39 | 100 | 97.62 |
| Average | 98.20 | 98.16 | 95.91 | 96.97 |

**Table 2** Classification metrics for SSD algorithm

| Class | Accuracy | Precision | Recall | F1-score |
|---|---|---|---|---|
| Voltage Source | 100 | 100 | 100 | 100 |
| Capacitor | 99.17 | 100 | 96.30 | 98.11 |
| Inductor | 98.34 | 97.29 | 92.31 | 94.73 |
| Resistor | 98.34 | 96.42 | 96.42 | 96.42 |
| Diode | 99.58 | 97.62 | 100 | 98.80 |
| Average | 98.62 | 98.27 | 97.01 | 97.62 |

**Table 3** Classification metrics for YOLOv3 algorithm

| Class | Accuracy | Precision | Recall | F1-score |
|---|---|---|---|---|
| Voltage Source | 100 | 100 | 100 | 100 |
| Capacitor | 98.34 | 96.30 | 96.30 | 96.30 |
| Inductor | 99.17 | 97.44 | 97.44 | 97.44 |
| Resistor | 99.17 | 96.55 | 100 | 98.24 |
| Diode | 100 | 100 | 100 | 100 |
| Average | 99.17 | 98.06 | 98.75 | 98.40 |

**Table 4** Classification metrics for YOLOv5 algorithm

| Backbone | Accuracy | Precision | Recall | F1-score |
|---|---|---|---|---|
| YOLOv5 | **99.17** | 98.06 | 98.75 | **98.40** |
| YOLOv3 | 98.62 | 98.27 | 97.01 | 97.62 |
| SSD300 | 98.20 | 98.16 | 95.91 | 96.97 |

**Table 5** Comparison of classification metrics

the last row of the matrix (in the column corresponding to the detected class).

## 4.4 Comparison and Discussion

The performance of three different object detection algorithms is compared in Table 1. Also, the performance of classification by these models are tabulated in Tables 2-4. These tables are extracted from the confusion matrices computed on respective models as shown in Figures 6-8. The trained models are evaluated on 51 different types of test images. Although mAP of YOLOv5 and YOLOv3 is the same, the time taken for execution of YOLOv5 is almost half of the YOLOv3. SSD model didn't perform well when compared to the YOLO algorithms. This might be due to the lower input image resolution given to the model compared to that of YOLO, but in any case, the time of execution increases when the input image size to SSD is increased. From Table 5, the best accuracy and F1-score for classification is obtained for the YOLOv5 algorithm. Hence, YOLOv5 is used to detect components in a given circuit owing to its low time of execution, better classification, and higher mAP. The shapes of resistor and inductor may look alike in some cases, hence compared to other components they have low average precision, whereas voltage source, capacitor and diode were detected with almost 100% AP since their shapes are quite different when compared to the rest of the components. The results of circuit recognition depend on both components detection and node detection. The component detection works for almost all circuits in the test set whereas node detection works well for most of the cases. Out of the 51 circuits in 47 circuits, nodes are detected accurately whereas in the remaining circuits one or two nodes are missed. As a whole, out of 51 hand-drawn circuits, the detection box or vice versa). If there are duplicates, the best match (greater IoU) is always selected.

4. The confusion matrix is updated accordingly to reflect the above matches.

5. Objects that are part of the ground-truth but weren't detected are counted in the last column of the matrix (in the row corresponding to the ground-truth class). Objects that were detected but aren't part of the confusion matrix are counted in above-proposed system is working well for 41 hand-drawn circuits. Hence in a majority of circuits, the above-proposed system works perfectly well.

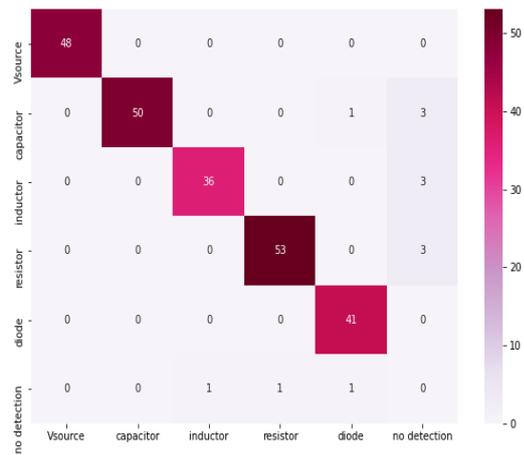

**Fig. 6** Confusion matrix for SSD algorithm

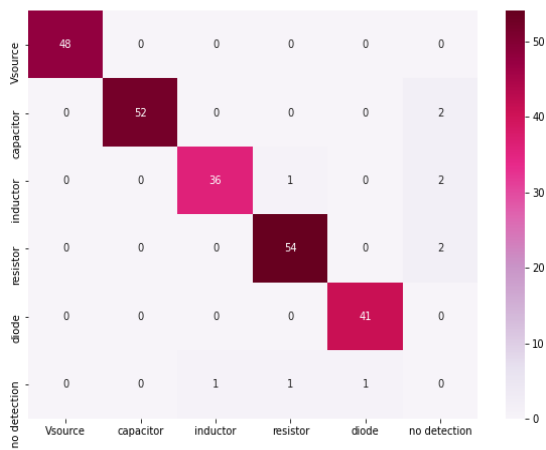

**Fig. 7** Confusion matrix for YOLOv3 algorithm

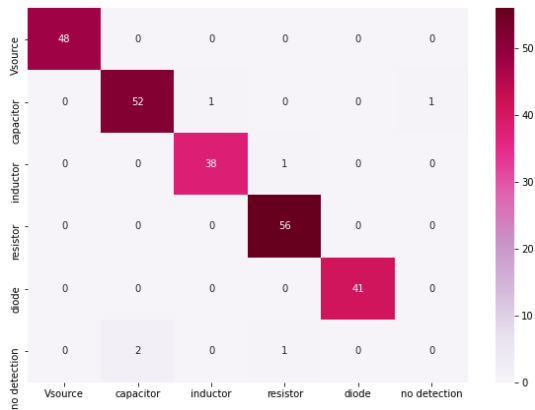

**Fig. 8** Confusion matrix for YOLOv5 algorithm

We have also evaluated the inference time complexity of the proposed approach. With YOLOv5 being light weight and real-time object detection approach and the classical image processing techniques employed in the proposed approach for node and terminal recognition being lower complexity algorithms, the expected time complexity of the proposed approach is minimal. The average inference time complexity on 20 images was found to be 0.33s,

| Method | Component recognition accuracy | Node recognition accuracy |
|---|---|---|
| Edwards et. al. [9] | 86 | 92 |
| Proposed method | 99 | 92 |

**Table 6** Comparison with the state-of-the-art approach

which makes the proposed approach useful for real-time applications.

Some limitations of the above-proposed system which will be addressed in future work are:

1. This algorithm works only if the branch contains one component.
2. Since, only 80% is the accuracy of training, it may fail to recognize some components, if it fails then the component is not separated from the input image, which disturbs the process of finding nodes. Hence, total recognition of the circuit fails.
3. Since, mapping is done based on distance, for some cases where the drawing of the electrical circuit is not as expected, there the recognition of the circuit fails.
4. The main challenges of handwriting are one cannot expect straight lines, the lines with same force, speed, paper, and shape. Also, there are related issues with the nature of images captured under varying lighting conditions and using various resolution cameras.

## 4.5 Comparison with existing methods

There are no approaches in the literature that have used an object detection algorithm to address the problem of circuit recognition, they were more focused on the classification of different components. The best approach in classification is by Dey et al [1] which used a dataset of various electronic components such as voltage source, capacitor, diode, logic gates, transistors etc. Whereas we have a dataset with electronic circuits, hence we cannot compare the architecture in [1] with our dataset. All the other approaches in the literature also use electronic component datasets to train the models and classification accuracy to evaluate the performance of their classifiers. To the best of our knowledge, the proposed approach in this paper is the first of its kind, which trains an object detection algorithm on a hand-drawn electrical circuit dataset. The proposed method is also a novel approach that addresses the problem of reconstructing the whole circuit. We have evaluated the reconstruction of the circuit based on component detection accuracy and node recognition accuracy. The best approach in component detection and node recognition [9] have also attempted the whole process of reconstruction of a circuit. Table 6 shows the comparison of our approach with [9]. However, the whole process proposed in [9] depends on thinning morphological operation, which isn't stable and, in most cases, when a circuit is thinned, more gaps are created which in turn affects the component and node recognition accuracy. On the other hand, the proposed model is independent of this morphological operation and is based on an object detection model to improve the component recognition accuracy.

## 5. Conclusion

Automatic generation of circuit schematics from hand drawn circuit diagrams can ease the workflow in electronic circuit design. A complete solution is presented to rebuild a hand drawn electrical circuit using image processing and object detection algorithms. A well-trained light weight deep learning model such as YOLO will give accurate classification and detection of components which can further help in tracing the connections and is more robust to varying lighting conditions and drawing styles when compared to classical image processing-based approaches. A novel Hough transform and k-means clustering method are employed to detect the nodes in a circuit. Using YOLOv5, 98.2% $mAP_{0.5}$ is achieved, and this can be used in real time applications. Also using the above proposed method accuracy of 92% is achieved for node recognition. All together a given hand drawn circuit can be reconstructed with an 80% accuracy using the proposed method with an inference time of nearly 0.33 seconds. A much more diverse dataset is required so that an end to end deep learning approach can be attempted in future, where even the node and terminal recognition can be done using neural networks.

## Declarations

**Conflicts of interest:** The authors declare that they have no conflict of interest.